\documentclass[11pt,a4paper]{article}
\usepackage{authblk}
\usepackage[hyperref]{eacl2021}
\usepackage{times}
\usepackage{latexsym}

\usepackage{microtype}

\usepackage{url}
\usepackage{graphicx}
\usepackage{amsmath}
\usepackage{amssymb}
\usepackage{pgfplots}
\pgfplotsset{compat=1.8}
\usepackage{mathtools}
\usepackage{tikz}
\usepackage{wrapfig}
\usepackage{caption}
\usepackage{subcaption}
\usepackage{cleveref}
\usepackage{tabularx}

\usepackage{algorithm}
\usepackage[noend]{algpseudocode}
\usepackage{relsize}
\usepackage{multirow}
\usetikzlibrary{fit,calc}
\usepackage{xcolor} 
\usepackage{pifont}
\usepackage{arydshln}

\usepackage{booktabs}

\usepackage{microtype}
\usepackage{array,etoolbox}

\aclfinalcopy 
\def\aclpaperid{33} 

\graphicspath{{./figures/}}

\title{Few-Shot Learning of an Interleaved Text Summarization Model by Pretraining with Synthetic Data}

\author[1,2]{\bf Sanjeev Kumar Karn}
\author[ ]{\bf Francine Chen}
\author[3]{\bf Yan-Ying Chen}
\author[4]{\bf Ulli Waltinger}
\author[1]{\bf Hinrich Sch\"{u}tze}
\affil[1]{Center for Information and Language Processing (CIS), LMU Munich}
\affil[2]{Machine Intelligence, Siemens Healthineers, Princeton}
\affil[3]{Toyota Research Institute, Los Altos, California}
\affil[4]{Machine Intelligence, Siemens AG, Munich}
\affil[ ]{\tt skarn@cis.lmu.de  francine@acm.org yan-ying.chen@tri.global }
\affil[ ]{\tt ulli.waltinger@siemens.com inquiries@cislmu.org}

\def\figlabel#1{\label{fig:#1}\label{p:#1}}
\def\figref#1{Figure~\ref{fig:#1}}
\def\eqref#1{Eq.~\ref{eqn:#1}}

\def\tabref#1{Table~\ref{tab:#1}}
\def\tablabel#1{\label{tab:#1}\label{p:#1}}

\definecolor{forestgreen}{rgb}{0.13, 0.55, 0.13}

\newcommand\boldblue[1]{\textcolor{blue}{\textbf{#1}}}
\newcommand\boldgreen[1]{\textcolor{forestgreen}{\textbf{#1}}}
\newcommand\boldred[1]{\textcolor{red}{\textbf{#1}}}

\newcommand\Tstrut{\rule{0pt}{2.6ex}}  
\newcounter{magicrownumbers}

\def\algref#1{Algorithm.~\ref{alg:#1}}
\algdef{SE}[VARIABLES]{Variables}{EndVariables}
   {\algorithmicvariables}
   {\algorithmicend\ \algorithmicvariables}
\algnewcommand{\algorithmicvariables}{\textbf{global}}

\begin{document}

\maketitle

\begin{abstract}
  Interleaved texts, where posts belonging to different threads occur in a sequence, commonly occur in online chat posts, so that it can be time-consuming to quickly obtain an overview of the discussions. Existing systems first disentangle the posts by threads and then extract summaries from those threads. A major issue with such systems is error propagation from the disentanglement component. While end-to-end trainable summarization system could obviate explicit disentanglement, such systems require a large amount of labeled data. To address this, we propose to pretrain an end-to-end trainable hierarchical encoder-decoder system using synthetic interleaved texts. We show that by fine-tuning on a real-world meeting dataset (AMI), such a system out-performs a traditional two-step system by 22\%. We also compare against transformer models and observed that pretraining with synthetic data both the encoder and decoder outperforms the BertSumExtAbs transformer model which pretrains only the encoder on a large dataset. 
\end{abstract}

\section{Introduction}
Interleaved texts are increasingly common, occurring in social media conversations such as Slack and Stack Exchange, 
where posts belonging to different threads may be intermixed in the post sequence; see a meeting transcript from the AMI corpus \cite{7529878bc1a143dbad4fa019e742fdb8} in \tabref{ami_sample}.
Due to the mixing, getting a quick sense of the different conversational threads 
is often difficult. 

In conversation disentanglement, interleaved posts are grouped by the thread. However, a reader still has to read all posts in each cluster of threads to get the gist. 
\newcite{P18-1062} proposed a two-step system that takes an interleaved text as input and first disentangles the posts thread-wise by clustering, and then compresses the thread-wise posts to single-sentence summaries. However, 
disentanglement e.g., \newcite{wang2009context}, propagates error to the downstream summarization task. 
An end-to-end supervised summarization system 
that implicitly identifies the conversations would eliminate error propagation. 
However, labeling of interleaved texts is a difficult and expensive task  \cite{barker2016sensei,aker2016automatic,verberne2018creating}.

\begin{table}[!ht]
\begin{center}
\resizebox{1.0\linewidth}{!}{
\begin{small}
\begin{tabular}{m{0.96\linewidth}}
\hline
\multicolumn{1}{c}{AMI Utteracnes}\\
\hline
\Tstrut\dots Who is gonna do a PowerPoint presentation ? Think we all Huh. You will \dots\\
\dots \textsuperscript{$\xi$}and uh the sender will send to the telly itself an \boldblue{infra-red signal} to tell it to switch on or switch\ldots\\
\dots \textsuperscript{$\zeta$}so y so it's so it's so you got so that's something we should have a look into then i when desi when \boldred{designing the ergonomics} of see have a look \ldots\\
\dots\textsuperscript{$\psi$},the little tiny weeny  \boldblue{batteries}, all like special long-lasting \boldblue{batteries}\ldots\\
\multicolumn{1}{c}{\dots}\\
\hline
\multicolumn{1}{c}{Summary}\\
\hline
\Tstrut1) the project manager had the team members re-introduce \ldots 
\\
2) the industrial designer discussed the interior workings of a remote and the team discussed \boldblue{options for batteries} and \boldblue{infra-red signals}.\\
\multicolumn{1}{c}{\dots}\\ \vspace{-0.5em}
5) the marketing expert presented research on consumer preferences on remotes in general and on voice recognition and the team discussed the option to have an \boldred{ergonomically designed} remote.\\
\multicolumn{1}{c}{\dots}\\
\hline
\end{tabular}
\end{small}
}
\end{center}
\caption{\tablabel{ami_sample} The top section shows AMI ASR transcripts and the bottom section shows human-written summaries. $\xi$=150\textsuperscript{th}, $\zeta$=522\textsuperscript{th} and $\psi$=570\textsuperscript{th} utterances. $a$) refer to the $a^{th}$ sentence in a multi-sentence summary.} 
\end{table}

We propose a pretraining approach to tackle these issues. We
synthesized a corpus of interleaved text-summary pairs out
of a corpus of regular document-summary pairs and train an
end-to-end trainable encoder-decoder system. To generate the
summary the model learns to infer (disentangle) the major
topics in several threads. We show on synthetic and
real-world data that the encoder-decoder system not only
obviates a disentanglement component but also enhances
performance. Thus, the summarization task acts as an
auxiliary task for the disentanglement. Additionally, we
show that fine-tuning of the encoder-decode system with the
learned disentanglement representations on a real-world AMI
dataset achieves a substantial increment in evaluation
metrics despite a small number of labels.

We also propose using a hierarchical attention in the encoder-decoder system with three levels of information from the interleaved text; posts, phrases, and words, rather than traditional two levels; post and word \cite{AAAI1714636,DBLP:conf/conll/NallapatiZSGX16,tan2017neural,cheng2016neural}.

The remaining paper is structured as follows.
In Section 2, we discuss related work. In Section 3,
we provide a detailed description of our hierarchical seq2seq model. In Section 4, we provide a detailed description on the synthetic data creation algorithm. In Section 5, we describe and discuss the experiments. And in Section 6, we present our conclusions. 

\section{Related Work}
\newcite{ma2012topic,aker2016automatic,P18-1062} each designed a system that summarizes posts in multi-party conversations in order to provide readers with overview on the discussed matters. They broadly follow the same two-step approach: cluster the posts and then extract a summary from each cluster.
 
There are two kinds of summarization: abstractive and extractive. In abstractive summarization, the model utilizes a corpus level vocabulary and generates novel sentences as the summary, while extractive models extract or rearrange the source words as the summary. Abstractive models based on neural sequence-to-sequence (seq2seq) \cite{DBLP:conf/emnlp/RushCW15} proved to generate summaries with higher ROUGE scores than the feature-based abstractive models.

\newcite{P15-1107} proposed an encoder-decoder (auto-encoder) model that utilizes a hierarchy of networks: word-to-word followed by sentence-to-sentence. Their model is better at capturing the underlying structure than a vanilla sequential encoder-decoder model (seq2seq). \newcite{krause2016paragraphs} and \newcite{P18-1240} showed multi-sentence captioning of an {\it image} through a hierarchical Recurrent Neural Network (RNN), topic-to-topic followed by word-to-word, is better than seq2seq. 
These works suggest a hierarchical decoder, thread-to-thread followed by word-to-word, may intrinsically disentangle the posts, and therefore, generate more appropriate summaries.

Integration of attention into a seq2seq model \cite{DBLP:journals/corr/BahdanauCB14} led to further advancement of abstractive summarization \cite{DBLP:conf/conll/NallapatiZSGX16,DBLP:conf/naacl/ChopraAR16}. \newcite{DBLP:conf/conll/NallapatiZSGX16} devised a hierarchical attention mechanism for a seq2seq model, where two levels of attention distributions over the source, i.e., sentence and word, are computed at every step of the word decoding. Based on the sentence attentions, the word attentions are rescaled. 
Our hierarchical attention is more intuitive, computes post(sentence)-level and phrase-level attentions for every new summary sentence, and is trained end-to-end.

Semi-supervised learning has recently gained popularity as it helps training parameters of large models without any training data. Researchers have pre-trained masked language models, in which only an encoder is used to reconstruct the text, e.g., BERT \cite{devlin2018bert}. \newcite{liu-lapata-2019-text} used BERT as seq2seq encoder and showed improved performance on several abstractive summarization tasks. Similarly, researchers have published pre-trained seq2seq models using a different semi-supervised learning technique, where a seq2seq model is learned to reconstruct the original text, e.g., BART \cite{lewis2019bart} and MASS \cite{song2019mass}. In this work, we rely on transfer learning and demonstrate that by pretraining with appropriate interleaved text data, a seq2seq model readily transfers to a new domain with just a few examples.      

\begin{figure*}[t!]
\centering
\includegraphics[width=0.90\textwidth]{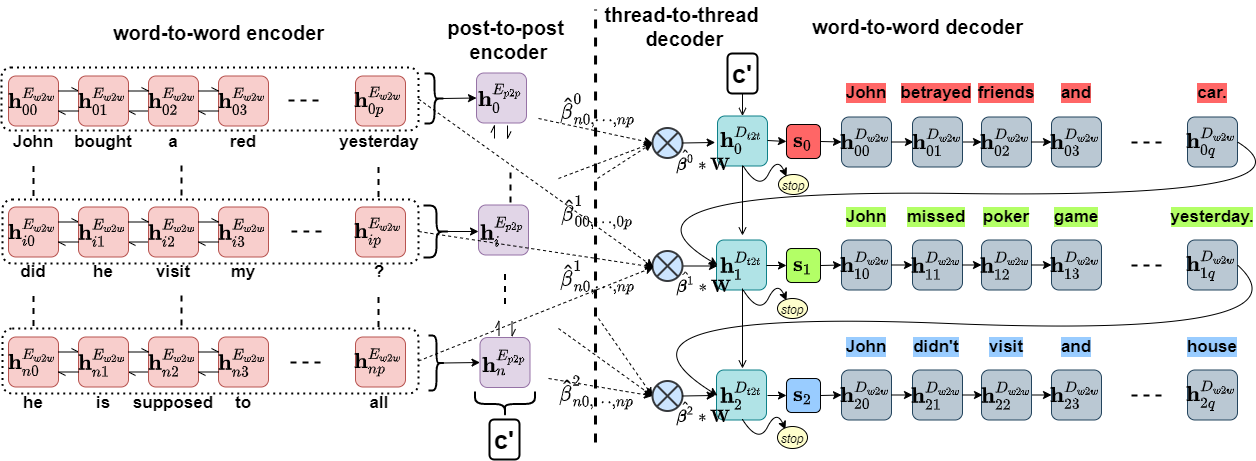}
\caption{Hierarchical encoder-decoder architecture. On the left, interleaved posts are encoded hierarchically, i.e., word-to-word ($E_{w2w}$) followed by post-to-post ($E_{p2p}$). On the right, summaries are generated hierarchically, thread-to-thread ($D_{t2t}$) followed by word-to-word ($D_{t2t}$). 
}\figlabel{architecture_joint}
\end{figure*}

\section{Model}
Our hierarchical encoder (see \figref{architecture_joint} left hand section) is based on \newcite{AAAI1714636}, where word-to-word and post-to-post encoders are bi-directional LSTMs. 
The word-to-word BiLSTM encoder ($E_{w2w}$) runs over word embeddings of post $\mathit{P}_i$ and generates a set of hidden representations, $\langle\mathbf{h}^{{E_{w2w}}}_{i,0},\ldots,\mathbf{h}^{{E_{w2w}}}_{i,p}\rangle$, of $d$ dimensions. 
The average pooled value of the word-to-word representations of post $\mathit{P}_i$ ($\frac{1}{p}\sum_{j=0}^{p} \mathbf{h}^{{E_{w2w}}}_{i,j}$) is input to the post-to-post BiLSTM encoder ($E_{t2t}$), which then generates a set of representations, $\langle\mathbf{h}^{E_{p2p}}_{0},\ldots,\mathbf{h}^{E_{p2p}}_{n}\rangle$, corresponding to the posts. 
Overall, for a given channel $\mathit{C}$, output representations of word-to-word, $\mathbf{W}$, and post-to-post, $\mathbf{P}$, has $n\times p\times 2d$ and $n\times 2d$ dimensions respectively.
The hierarchical decoder has two uni-directional LSTM decoders, thread-to-thread and word-to-word (see right-hand side in \figref{architecture_joint}). 

At step $k$ of thread decoder (${D_{t2t}}$), we compute elements of post-level attention as $\gamma^{k}_{i} = \sigma(\mbox{attn}^{\gamma}(\mathbf{h}^{D_{t2t}}_{k-1}, \mathbf{P}_{i})\enskip i \in \{1,\dotsc,n\}
$, where $\mbox{attn}^{\gamma}$ aligns the current thread decoder state vector $\mathbf{h}^{D_{t2t}}_{k-1}$ to vectors of matrix $\mathbf{P}_{i}$. 
A phrase is a short sequences of words in a sentence/post. Phrases in interleaved texts are equivalent to visual patterns in images, and therefore, attending phrases are more relevant for thread recognition than attending posts. Thus, we have phrase-level attentions focusing on words in a channel and with a responsibility of disentangling threads. At step $k$ of thread decoder, we also compute a sequence of attention weights, $\boldsymbol{\beta^{k}}$ = $\langle\mathit{\beta}^{k}_{0,0},\ldots,\mathit{\beta}^{k}_{n,p}\rangle$, corresponding to the set of encoded word representations, $\langle\mathbf{h}^{w2w}_{0,0},\ldots,\mathbf{h}^{w2w}_{n,p}\rangle$, as
$\beta^{k}_{i,j}=\sigma(\mbox{attn}^{\beta}(\mathbf{h}^{D_{t2t}}_{k-1}, \mathbf{a}_{i,j})) \enskip \text{where} \enskip \mathbf{a}_{i,j} = add(\mathbf{W}_{i,j}, \mathbf{P}_{i}), \enskip i \in \{1,\dotsc,n\},\enskip j \in \{1,\dotsc,p\}
$. ${add}$ aligns a post representation to its word representations and does element-wise addition, and $\mbox{attn}^{\beta}$ maps the current thread decoder state $\mathbf{h}^{D_{t2t}}_{k-1}$ and vector $\mathbf{a}_{i,j}$ to a scalar value. 
Then, we use the post-level attention, $\boldsymbol{\gamma}^{k}$, to rescale the sequence of attention weights $\boldsymbol{\beta^{k}}$ to obtain phrase-level attentions $\boldsymbol{\hat{\beta}^{k}}$ as $\hat{\beta}^{k}_{i,j} = \beta^{k}_{i,j}*\gamma^{k}_{i}$.

A weighted representation of the words (crossed blue circle), $\sum_{i=1}^{n}\sum_{j=1}^{p}\hat{\beta}^{k}_{i,j}\mathbf{W}_{ij}$, is used as an input to compute the next state of the thread-to-thread decoder, $D_{t2t}$. Additionally, we also use the last hidden state $\mathbf{h}^{D_{w2w}}_{k-1,q}$ of the word-to-word decoder LSTM (${D_{w2w}}$) of the previously generated summary sentence as the second input to $D_{t2t}$. The motivation is to provide information about the previous sentence. 

The current state $\mathbf{h}^{D_{t2t}}_{k}$ is passed through a single layer feedforward network and a distribution over STOP=1 and CONTINUE=0 is computed: $\mathit{p}_{k}^{STOP} = \sigma(\mbox{g}\left({\mathbf{h}^{D_{t2t}}_{k}}\right)) $, where $\mbox{g}$ is a feedforward network. In \figref{architecture_joint}, the process is depicted by a yellow circle. The thread-to-thread decoder keeps decoding until  $\mathit{p}_{k}^{STOP}$ is greater than 0.5.

Additionally, the new state $\mathbf{h}^{D_{t2t}}_{k}$ and inputs to $D_{t2t}$ at that step are passed through a two-layer feedforward network, \text{r}, followed by a dropout layer 
to compute the thread representation $\mathbf{s}_k$.
Given a thread representation $\mathbf{s}_k$, the word-to-word decoder, a unidirectional attentional LSTM (${D_{w2w}}$), generates a summary for the thread; see the right-hand side of \figref{architecture_joint}. 
Our word-to-word decoder is based on \newcite{DBLP:journals/corr/BahdanauCB14}.

At step $l$ of word-to-word decoding of summary of thread $k$, we compute elements of word level attention, i.e., $\boldsymbol{\alpha}^{k,l}_{i,\cdot}$; we refer to \newcite{DBLP:journals/corr/BahdanauCB14} for further details on it. However, we use phrase-level word attentions for rescaling the word level attention as $\hat{\alpha}^{k,l}_{i,j} = norm(\hat{\beta}^{k}_{i,j} \times \alpha^{k,l}_{ij})$, where $norm$ (softmax) renormalizes the values. Thus, contrary to popular two-level hierarchical attention \cite{DBLP:conf/conll/NallapatiZSGX16,cheng2016neural,tan2017neural}, we have three levels of hierarchical attention and each with its responsibility and is coordinated through the rescaling operation. 

We train our model end-to-end to minimize the following objective: ${\underset{k=1}{\overset{m}{\sum}}}{\underset{l=1}{\overset{q}{\sum}}}\log{p}_\theta\left(y^{k,l}\vert\textit{w}_{k,\cdot<l}, \textbf{W}\right)+\lambda {\underset{k=1}{\overset{m}{\sum}}}\textit{y}_{k}^{STOP}\log(p_{k}^{STOP})$, 
where $\langle \mathit{w}^{k,0},\ldots, \mathit{w}^{k,q}\rangle$ and $\langle \mathit{y}^{k,0},\ldots, \mathit{y}^{k,q} \rangle$ are words in a ground-truth summary and $D_{w2w}$ generation respectively.
\begin{table}[!t]
\begin{center}
\resizebox{1.0\linewidth}{!}{
\begin{small}
\begin{tabular}{p{0.14\linewidth}p{0.88\linewidth}}
\hline
\multicolumn{2}{c}{\bf Notations}\\
\hline
\Tstrut$\textit{C}$ & A sequence of pairs of single-thread texts and single-sentence summaries\\
$a$& A minimum number of threads\\
$b$& A maximum number of threads\\
$m$& A minimum number of posts\\
$n$& A maximum number of posts\\
$\textsc{window}$&returns an iterator object that traverse a given sequence elements in windowed manner\\
$w$& window size\\
$t$& step size\\
$\textit{O}$ & An iterator object that returns a window size sequence of pairs of single-thread texts and single-sentence summaries\\
$\textit{E}$ & A window size sequence of pairs of single-thread texts and single-sentence summaries\\
$\textit{I}^\prime$ & A sequence of sentences\\
$\textit{M}^\prime$ & A sequence of single-sentence summaries\\
$\mathcal{U}$& A uniform sampling function\\
$\textit{A}$ & A single thread text as a sequence of sentences\\
$\textit{T}$ & A single-sentence summary as a sequence of words\\
$\hat{\textit{I}}$ & A multi-thread interleaved text\\
$\hat{\textit{M}}$ & A multi-sentence summary of an interleaved text\\
$\lbrace a_{\times b}\rbrace$& A set in which variable $a$ is repeated $b$ times\\
${\textit{S}}$&A sequence of positive integers\\
$\textsc{Reverse}$& reverses an array\\
$\textsc{Pop}$& removes the last element from an array\\
$\textit{P}$ & A sentence/post as a sequence of words\\
$\left[\cdot\right]$& array indexing operation\\
$X\setminus Y$& A set of elements that belong to $X$ but not to $Y$\\
$\textit{Z}$ & A sequence of pairs of multi-thread interleaved texts and multi-sentence summaries\\
\hline
\end{tabular}
\end{small}
}
\end{center}
\caption{\tablabel{algo_notations} We use lowercase italics for variables, uppercase italics for sets and sequences, math symbols for mathematical operations and uppercase words for methods.} 
\end{table}

\section{Synthetic Dataset}
Obtaining labeled training data for interleaved conversation
summarization is challenging. The available ones are either
extractive \newcite{verberne2018creating} or too small
\cite{barker2016sensei,anguera2012speaker} to train a neural
model and thoroughly verify the architecture. To get around
this issue, we synthesized a dataset by utilizing a corpus
of conventional texts for which summaries are available. We
created a corpus of interleaved texts from the abstracts and
titles of articles from the PubMed corpus
\cite{I17-2052}. We chose PubMed abstracts as it has, in contrast to
other corpora such as news articles or StackOverflow posts,
a single-sentence summary that can only be
comprehended out of a whole abstract. Further, the number of
sentences more closely resembles that of a conversationalist in
a conversation.

Random interleaving of the sentences from a small number of PubMed abstracts roughly resembles interleaved texts, and, correspondingly, interleaving of titles resembles its multi-sentence summary. We devised an algorithm for creating synthetic interleaved texts based on this idea. 

\textbf{Interleave Algorithm:}
The Interleave Algorithm generates interleaved texts, each
containing randomly interleaved sentences from a small
number of abstracts, where the number is a random value
within a specified range. The number of sentences used per
abstract is also a random value within a specified
range. Abstracts to be included in an interleaved text are first selected, then the selected abstracts are interleaved, and finally the interleaved texts together with a concatenation of the titles of the selected abstracts are returned.

\label{app:algo}
\begin{algorithm}[!t]
\begin{small}
\caption{Interleaving Algorithm}\label{alg:interleave_algo}
\begin{algorithmic}[1]
\Procedure{Interleave}{$\textit{C}, \textit{a}, \textit{b}, \textit{m}, \textit{n}$}
\State $\textit{O}, \textit{Z}\gets \textsc{window}(\textit{C}, w, t)$, Array()
\While {$\textit{O} \neq \emptyset$}
	\State $\textit{E}, \textit{I}^\prime, \textit{M}^\prime, \textit{S} \gets \textit{O}.\textsc{Next}()$, Array(), Array(), $\{\}$
	\State ${\textit{r}}$ $\sim$ $\mathcal{U}$($\textit{a}, \textit{b}$)	
	\For {$\textit{j}$ = 1 to $\textit{r}$} \Comment{Selection}
		\State $\textit{A}, \textit{T}  \gets \textit{E}[\textit{j}]$
		\State ${\textit{q}}$ $\sim$ $\mathcal{U}$($\textit{m}, \textit{n}$)
		\State $\textit{I}^\prime.\textsc{Add}(\textit{A}$[1:$\textit{q}$])
		\State $\textit{M}^\prime.\textsc{Add}(\textit{T}$)
		\State $\textit{S} \gets \textit{S}\cup\lbrace j_{\times q}\rbrace$
\EndFor
\State $\hat{\textit{I}}, \hat{\textit{M}}, \textit{l} \gets$ Array(), Array(), $|{\textit{S}}|$	
	\For {1 to $\textit{l}$}	\Comment{Interleaving}
		\State $\textit{k} \gets \mathcal{U}(\textit{S})$
		\State $\textit{P} \gets \textsc{Reverse}(\textit{I}^\prime$[$\textit{k}]).\textsc{pop}()$
		\State $\hat{\textit{I}}.\textsc{Add}(\textit{P}$)		
		\State $\textit{T} \gets \textit{M}^{\prime}$[$\textit{k}$]
		\If {$\textit{T} \not\in \hat{\textit{M}}$}:		
			\State $\hat{\textit{M}}.\textsc{Add}(\textit{T}$)
		\EndIf
		\State $\textit{S} \gets \textit{S}\setminus{\textit{k}}$
	\EndFor
	\State $\textit{Z}.\textsc{Add}(\hat{\textit{I}}$;$\hat{\textit{M}}$)
\EndWhile
\State \Return $\textit{Z}$
\EndProcedure
\end{algorithmic}
\end{small}
\end{algorithm}

We first refer to \tabref{algo_notations} for terms and
notations used in \algref{interleave_algo}.
$\textsc{Interleave}$ takes a corpus of abstract-title
pairs, $\textit{C}$ =
$\langle\textit{A}_1;\textit{T}_1,\textit{A}_{2};\textit{T}_{2},
\ldots,\textit{A}_{N};\textit{T}_{N}\rangle$, and returns a
sequence of pairs of multi-thread interleaved texts and multi-sentence summaries, $Z$. Each
interleaved text in the generated sequence will contain a number of threads
ranging between $a$ to $b$, where the number is randomly
selected. Each thread, in turn, will contain a number of
posts or sentences ranging between $m$ and $n$, where this number is also
randomly selected. The $\textsc{window}$ function is given $C$, a
desired window size, $w$, and step size, $t$, and it returns an iterator object,
$\textit{O}$, of size
$\frac{|\mathit{N}|-\textit{w}}{t}+1$. $\textsc{window}$
helps to enlarge the interleaved corpus without redundancy
as abstracts are randomly sampled out of an iterator element, $\textit{E}$, and also new abstracts are always included in the next element through sliding. Similarly, sets of sentences are randomly sampled out of the selected abstracts. Thus, interleaved text-summary pairs in the corpus are different. The two parts of the $\textsc{Interleave}$ algorithm, Selection and Interleaving, will be described next. 

\textbf{Selection:} $\mathcal{U}$ in step 5 determines the number of threads, $r$. Then, thread candidates for an interleaved text are chosen out of an iterator element $\textit{E}$, a window size sequence of pairs of single-thread texts and single-sentence summaries. Next, post candidates for each selected thread, $\textit{A}$, are chosen. $\mathcal{U}$ in step 8 determines the number of posts, $q$. Thread indices are repeated as many times as its posts and stored in a set, $\textit{S}$. 

\textbf{Interleaving:} In every step in a loop of a size equivalent to the length of indices $\textit{S}$, $\mathcal{U}$ randomly selects a thread index. $\textsc{Reverse}$ and $\textsc{Pop}$ in step 15 help in selecting a post, $P$, in the selected thread in a FIFO manner. The single-sentence summary, $\textit{T}$, of the thread is added to the multi-sentence summary sequence, $\hat{\textit{M}}$, if it didn't exist previously. 

As an interleaved text-summary pair in the corpus has a thread size between \textit{a} and \textit{b} and post size per thread between \textit{m} and \textit{n}, the larger the difference between \textit{a} and \textit{b} and \textit{m} and \textit{n} in a corpus, the harder the disentangling and summarization task. So, we vary these 
parameters and create different synthetic corpora of varying
difficulty for the experiments. 
\tabref{ilv_example} shows an example of a data instance from a Interleaved PubMed corpus compiled using $a$=2, $b$=5, $m$=2 and $n$=5.

\begin{table}[!t]
\begin{center}
\resizebox{0.95\linewidth}{!}{
\begin{small}
\begin{tabular}{m{0.99\linewidth}}
\hline
\multicolumn{1}{c}{Interleaved text}\\
\hline
\Tstrut{\raise0.8ex\hbox{$\boldsymbol\pi$}} \ldots conducted to evaluate the influence of excessive sweating during long-distance running on the urinary concentration of caffeine\ldots\\
\Tstrut{\raise0.8ex\hbox{$\boldsymbol\omega$}} \ldots to assess the effect of a program of supervised fitness walking and patient education on functional status\ldots\\
\multicolumn{1}{c}{\dots}\\
\Tstrut{\raise0.8ex\hbox{$\boldsymbol\pi$}} \ldots 102 patients with a documented diagnosis of primary osteoarthritis of one or both knees participated\ldots\\
\Tstrut{\raise0.8ex\hbox{$\boldsymbol\phi$}} \ldots examined the effects of intensity of training on ratings of perceived exertion (\ldots\\
\multicolumn{1}{c}{\dots}\\ 
\hline
\multicolumn{1}{c}{Summary}\\
\hline
\Tstrut \raise0.8ex\hbox{$\boldsymbol\pi$} caffeine in sport. influence of endurance exercise on the urinary caffeine concentration.\\
\Tstrut \raise0.8ex\hbox{$\boldsymbol\omega$} supervised fitness walking in patients with osteoarthritis of the knee. a randomized , controlled trial.\\
\Tstrut \raise0.8ex\hbox{$\boldsymbol\phi$} the effect of training intensity on ratings of perceived exertion.\\
\hline
\end{tabular}
\end{small}
}
\end{center}
\caption{
\tablabel{ilv_example}An example of a synthetic Interleaved text and summary pair compiled using PubMed corpus and \algref{interleave_algo}. It includes three threads (abstracts) identifiable through superscribed symbols $\boldsymbol\pi$, $\boldsymbol\omega$, and $\boldsymbol\phi$.}
\end{table}
\section{Experiments}

\textbf{Parameters:}
For the word-to-word encoder, the steps are limited to 20, while the steps in the word-to-word decoder are limited to 15. The steps in the post-to-post encoder and thread-to-thread decoder depend on the corpus type, e.g., a Hard corpus compiled using $a$=2, $b$=5, $m$=2 and $n$=5 has 25 steps in post-to-post encoder, i.e., \textit{b}$\times$\textit{n} (the maximum possible size of posts in an item in the corpus) and 5 steps in thread-to-thread decoder, i.e., \textit{b} (the maximum possible threads in an item in the corpus). We initialized all weights, including word embeddings, with a random normal distribution with mean 0 and standard deviation 0.1. The embedding vectors and hidden states of the encoder and decoder in the models are set to dimension 100. Texts are lowercased. The vocabulary size is limited to 8000. We pad short sequences with a special token, $\langle PAD\rangle$. We use Adam \cite{DBLP:journals/corr/KingmaB14} with an initial learning rate of .0001 and batch size of 64 for training. 
The training, evaluation and test sets in a Hard Interleaved PubMed corpus ($a$=2, $b$=5, $m$=2 and $n$=5) are of sizes of 170k, 4k and 4k respectively. 

We report ROUGE-1, ROUGE-2, and ROUGE-L as the quantitative evaluation of the models.

\def\evalsep{0.20cm}
\def\perlevelsep{0.125cm}
\def\perlevelsepBase{0.08cm}

\begin{table}[!t]
      \centering
        \begin{center}
        \resizebox{1.0\linewidth}{!}{
\begin{tabular}
{
l@{\hspace{\perlevelsepBase}}|l|c@{\hspace{\perlevelsep}}c@{\hspace{\perlevelsep}}c|}
{Input Text} &{Model} & Rouge-1 &  Rouge-2 &  Rouge-L\\ 
\hline
dis (upper bd) & hier2hier & \bf39.09& \bf30.11& \bf15.22\\
\hline
\multicolumn{2}{c|}{\cite{P18-1062}}& 29.11& 15.76& 10.13\\
ent& hier2hier & \bf37.11& \bf27.97& \bf14.26\\
\end{tabular}
}
        \end{center}
        \caption{
        \tablabel{base_perf_comp} Synthetic interleaved text summarization performance (Rouge Recall-Scores) comparing models when the threads are disentangled (top section, upper bound) and when the threads are entangled (bottom section, real-world) on an Interleaved PubMed Corpus. \textbf{dis} = disentangled (ground-truth) and \textbf{ent} = entangled.  
}
\end{table}

\textbf{Upper-bound:} 
In upper-bound experiments, we check the impact of disentanglement on the abstractive summarization models. 
In order to do this, we first evaluate the performance of a model when provided the ground-truth disentanglement (thread indices) information. We also evaluate the performance of models for either end-to-end or two-step summarization. 

\textbf{Ground-truth Disentangled:} The ground-truth disentanglement information is used and posts of threads are disentangled and concatenated (posts are thread-wise sequentially arranged, i.e., non-interleaved). The first row in \tabref{base_perf_comp} shows performance of the hier2hier summarization model. Clearly, the model can easily detect a thread boundary in concatenated threads and perform very well
, and therefore, sets an upper bound for the task.

\textbf{No disentanglement:} In real-world scenarios, i.e., with no disentanglement, \newcite{P18-1062}'s unsupervised two-step system first disentangles/clusters the posts thread-wise and then compresses clusters to single-sentence summaries. While hier2hier is trained end-to-end, and therefore, generates multi-sentence summaries for a given interleaved text. \tabref{base_perf_comp} shows \newcite{P18-1062} performs worse than hier2hier (compare rows 2 and 3), indicating that a hier2hier model trained on a sufficiently large dataset is better at summarization than the unsupervised sentence compression method, especially in fluency as indicated by an approximately 12 point increase in Rouge-2. Additionally, the hier2hier model trained on entangled texts achieves slightly lower performance to when it is trained on disentangled texts (compare rows 1 and 3), indicating that the disentanglement component can be avoided if summaries are available. The bottom section in \tabref{result_example} show an example of the model generations (shown in color). The top indexes of the phrase-level attention ($\boldsymbol{\hat{\beta}}$) is directly visualized in the table through the color coding matching the generation. This shows phrase level attention actually supports in learning to disentangle the interleaved texts.
\begin{table}[!t]
\begin{center}
\resizebox{0.95\linewidth}{!}{
\begin{small}
\begin{tabular}{m{0.99\linewidth}}
\hline
\multicolumn{1}{c}{Interleaved text}\\
\hline
\Tstrut \ldots conducted \boldblue{to evaluate the influence of} excessive \boldblue{sweating} during \boldblue{long-distance} running \boldblue{on} the urinary concentration of \boldblue{caffeine}\ldots\\
\Tstrut \ldots \boldgreen{to assess the effect of} a \boldgreen{program of} supervised \boldgreen{fitness} walking and patient education on functional status\ldots\\
\multicolumn{1}{c}{\dots}\\
\Tstrut \ldots 102 patients with a documented diagnosis of primary osteoarthritis of one or both knees participated\ldots\\
\Tstrut \ldots \boldred{examined the effects of intensity of training} on ratings of perceived exertion \boldred{(}\ldots\\
\multicolumn{1}{c}{\dots}\\
\hline
\multicolumn{1}{c}{Generation}\\
\hline
\Tstrut \boldblue{effect of excessive [UNK] during [UNK] running on the urinary concentration of caffeine .}\\
\Tstrut \boldgreen{effect of a physical fitness walking on functional status , pain , and pain}\\
\Tstrut \boldred{effects of intensity of training on perceived [UNK] in [UNK] athletes .}\\
\hline
\end{tabular}
\end{small}
}
\end{center}
\caption{
\tablabel{result_example}An example of hier2hier generated summary sentences of a three thread interleaved text. Summaries are coloured differently and colors of attended phrases ($\boldsymbol{\beta}$) in the text are identical to those of the generations. The table is best viewed in color.}
\end{table}

\begin{figure}[t!]
\centering
\includegraphics[width=0.45\textwidth]{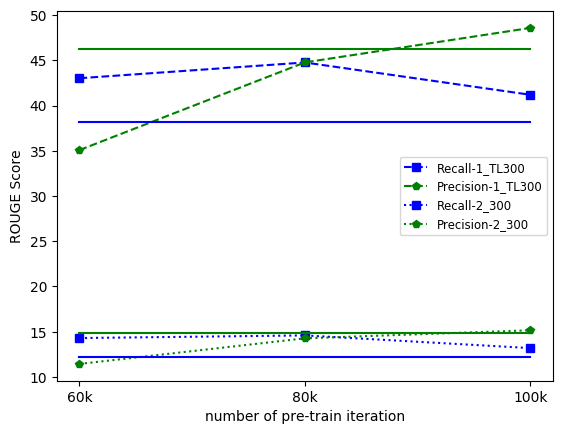}
\caption{ROUGE uni- and bi-gram precision (green) and recall (blue) of AMI fine-tuned hier2hier models with different numbers of pretraining iterations. Maximum words in a summary is 300. As a reference, solid horizontal lines show the scores of a model trained only on AMI.
}\figlabel{tlearn_plot}
\end{figure}
\textbf{Transfer Learning:} 
We utilize our interleaving algorithm and PubMed data to compile an interleaved corpus with a similar thread distribution as a corpus of real meetings, the AMI meeting corpus. AMI is a very small size corpus, so we have a train, eval and test split of 112, 10 and 20 respectively. Our analysis of the AMI corpus show that 90\% of meetings have $\leq$ 12 summary sentences while 60\% of meetings have $\geq$ 8 summary sentences, so we used 8 and 12 as the min (\textit{a}) and max (\textit{b}) number of threads respectively in the algorithm and create a synthetic corpus. We pretrain the hier2hier model for several iterations on the synthetic corpus, and then transfer and fine-tune the model on the AMI corpus with all parameters fixed except for the word-to-word decoder and hierarchical attention parameters. As PubMed and AMI are from different domains, we use the byte pair encoding (BPE) \cite{sennrich-etal-2016-neural} based subword dictionary. As shown in \tabref{ami_sota}, hier2hier readily transfers its disentangling knowledge, and therefore, obtains a boost in recall while maintaining its precision. The \newcite{li-etal-2019-keep} system has the best ROUGE-1 scores, however their model is not directly comparable as unlike \newcite{P18-1062} and our text-based model, it uses audio and video in addition to text. 

Additionally, we also performed transfer learning experiments with models pre-trained for a different number of iterations, and as seen in \figref{tlearn_plot}, hier2hier readily transfers its disentangling knowledge, and therefore, obtains a boost in recall while maintaining its precision. However, longer pretraining drives the model to generate shorter summaries similar to PubMed abstracts, and thereby, results in increasing precision and decreasing recall. 

We also experimented with state of the art transformer-based seq2seq models, e.g., BertSumExtAbs \cite{liu-lapata-2019-text} and BART \cite{lewis2019bart}. BertSumExtAbs requires fine-tuning of the encoder and a de novo training of decoder while both encoder and decoder of BART are only fine-tuned. We use only AMI data for the de novo training and fine-tuning purpose, and the bottom two rows in \tabref{ami_sota} show the results from these models.\footnote{Due to the small AMI data size, batch size and initial learning rate of BERTSumExt are set to 8 and 5e-4 respectively, batch size in BERTSumExtAbs is 16 and initial learning rates of BERT and transformer decoder in BERTSumExtAbs are 0.001 and 0.01 respectively.} Although our hier2hier\_t-learn also only requires fine-tuning of the decoder and hierarchical attention, a highly-sophisticated semi-supervised training of both the encoder and decoder of BART and larger model size (100x) yields better performance. However, for applications that have limited memory, as on some mobile devices, our model may be more desirable. Furthermore, despite a pre-trained encoder of BertSumExtAbs, a de novo training of a large size decoder with a tiny AMI data lead to over-fitting, and therefore, lower scores.

\textbf{Human Evaluation:} We also performed a qualitative evaluation of our system using human judgments. Following \newcite{chen-bansal-2018-fast}, we performed a comparative evaluation, where we provided six human judges (graduate students fluent in English) with meetings ($\approx 6000$ words) and summaries from three sources, i.e., human reference, two-step baseline  and hier2hier\_t-learn (here after referred to as the "our model"), and asked them to rate on a scale of 1 to 5 the two questions: 1) is the summary concise, fluent and grammatical (\textbf{fluency}) and 2) does the summary retain key information from the meeting (\textbf{relevancy})? 

\begin{table}[!t]
\begin{center}
\resizebox{1.\linewidth}{!}{
\begin{tabular}
{l|
c@{\hspace{\perlevelsep}}c@{\hspace{\perlevelsep}}c|
c@{\hspace{\perlevelsep}}c@{\hspace{\perlevelsep}}c|}
\Tstrut &\multicolumn{3}{c|}{\bf ROUGE-1}&\multicolumn{3}{c|}{\bf ROUGE-2}\\
Model &P &R &F1 &P &R &F1\\
\hline 
two-step \cite{P18-1062} &34.61 &41.84 &37.37 &6.92 &8.29 &7.45\\
hier2hier &46.30 &38.17 &41.30 &14.84 &12.23 &13.13\\
hier2hier\_t-learn &\bf{47.68} &\bf{44.37} &\bf{45.56} &\bf{16.02} &\bf{14.98} &\bf15.35\\
\hline
\cite{li-etal-2019-keep} &- &- &\bf53.29 &- &- &13.51\\
BertSumExtAbs &\bf55.95 &36.21 &43.24 &\bf18.35 &12.16 & 14.39\\
BART(base) &42.17 &\bf59.19 &49.03 &16.52 &\bf23.05 &\bf19.15\\
\end{tabular}
}
\end{center}
\caption{
\tablabel{ami_sota} Rouge Scores for summary size 300 words on the AMI Corpus. t-learn=transfer-leaning.  BART(base) = BART with 6 encoder and decoder layers and 140M parameters.  Li et al. uses audio+video in addition to text and the transformer models (the bottom two rows) have lots of extra data for pre-training.}
\end{table}

\begin{table}[!t]
\begin{center}
\resizebox{1.0\linewidth}{!}{
\begin{small}
\begin{tabular}{m{0.96\linewidth}}
\hline
\multicolumn{1}{c}{h2h\_t-learn summaries}\\
\hline
\Tstrut$1$) the project manager opened the meeting and went over the minutes of the previous\ldots\\
\multicolumn{1}{c}{\dots}\\
$3$) the industrial designer discussed the interior workings of a remote and the team\ldots\\
\multicolumn{1}{c}{\dots}\\
$5$) the group discussed the shape of the device and decided to make the device easier\ldots\\
\multicolumn{1}{c}{\dots}\\
\hline
\multicolumn{1}{c}{two-step \cite{P18-1062} summaries}\\
\hline
\Tstrut$1$) marketing report uh we observed remote control users in a usability lab\\
\multicolumn{1}{c}{\dots}\\
$7$) majority except for the forty five to fifty five year olds for some reason didnt want a voice act speech\ldots
\\
\multicolumn{1}{c}{\dots}\\
$14$) headed towards like a b a big yellow and black remote as far as maybe thats our next meeting that we discuss that\\
\multicolumn{1}{c}{\dots}\\
\hline
\end{tabular}
\end{small}
}
\end{center}
\caption{\tablabel{ami_model} The top and bottom sections show our hierarchical and the \newcite{P18-1062} system summaries respectively for ASR transcripts in \tabref{ami_sample}. $a$) refer to the $a^{th}$ sentence in a multi-sentence summary.}
\end{table}

We sampled six meetings (each with three summaries corresponding to three sources), duplicated them, and then randomly sampled two dissimilar meetings and assigned them to each judge to annotate. For reference, an annotation sample would be an ASR transcript and human written summaries as in \tabref{ami_sample} and our model and \newcite{P18-1062} summaries as in \tabref{ami_model}. The judges were not shown the source of the summaries. The twelve ratings that we received are converted into two binary comparisons and are summarized in \tabref{quali_eval}. 
Our model summaries were often judged to be better than the \newcite{P18-1062} system summaries in both fluency and relevancy. Gwet's AC1 and Brennan's and Prediger's kappa inter-rater agreement statistics show strong agreement for fluency. \footnote{Gwet's AC(1) and Brennan and Prediger's Kappa adjust the impact of the empirical distributions over the chance agreement, and therefore, are better suited for cases where the proportion of agreements on one class differs from that of another.} However, compared to human summaries, our model summaries were similar in terms of fluency but were lower in terms of relevancy, with inter-rater statistics indicating fair strength of agreement.

\begin{table}[t!]
\begin{center}
\resizebox{.99\linewidth}{!}{
\begin{tabular}{|p{1.5cm}|p{0.9cm}|p{0.9cm}|p{0.9cm}|p{0.9cm}|p{0.9cm}|}
\hline
\textbf{Metric} & \textbf{Win} & \textbf{Tie} & \textbf{Lose} & \textbf{gwet ac1}&\textbf{bp}\\
 \hline
 \multicolumn{6}{|c|}{Our Model vs. \newcite{P18-1062}} \\
\hline
Fluency &9 &2 &1 & 0.72 & 0.63\\
Relevancy  &9 &1 &2 & -0.02&-1.3\\  
 \hline
 \multicolumn{6}{|c|}{Our Model vs. Human Reference} \\
\hline
Fluency &3 &2 & 7 & 0.27 & 0.25\\
Relevancy  &2 &0 &10& 0.38& 0.25 \\  
\hline
\end{tabular}
}
\end{center}
\caption{\tablabel{quali_eval} Comparative ratings by human judges of summaries on fluency and relevancy metrics. gwet ac1 and bp refer to Gwet's AC(1) and Brennan-Prediger Kappa coefficients respectively.}
\end{table}

We also compared statistics of reference summaries against Our and \newcite{P18-1062} model generated summaries of maximum 300 words. 
We observe our model generates approximately 145 words outputs, which is close to ground-truth human written summaries of size approximately 165 words. However, the \newcite{P18-1062} system generates summaries of average 290 words. Further, the median number of threads (number of summaries) of our model, human written summaries and \newcite{P18-1062} are 8, 8.5, 17, respectively. This indicates our model is learning to generate  human-like summaries, while \newcite{P18-1062} aims to distill words up to the permissible limit, and therefore, has high recall and very low precision; see \tabref{ami_sota}. Additionally, our model has twice the \newcite{P18-1062} Rouge-2 values, which indicates high readability and was supported by human judges. Further, the difference in number of threads (summaries) between our model and reference are 
$\leq$ 3, 2, and 1 for 85\%, 65\%, and 40\% of cases, respectively.
This clearly indicates the strength of our hierarchical model in disentangling threads. 

\section{Conclusion}
We 
investigated the use of an end-to-end hierarchical encoder-decoder model,  hier2hier, with three levels of hierarchical attention for jointly summarizing and implicitly disentangling 
interleaved text. To train this model, we examined the use of pretraining using synthesized data and fine-tuning for adaptation to a new domain with limited labeled real-world data. 
On real-world AMI data, our fine-tuned end-to-end system outperforms a two-step system by 22\% (Rouge-1). Experiments were also conducted against the transformer-based BertSumExtAbs and BART systems, which indicate that these transformer models can also summarize interleaved texts. Specifically our hier2hier model also outperformed the transformer-based BertSumExtAbs but not BART, which suggests that use of pretraining of both the decoder as well as the encoder is important, and also indicates the utility of our synthetic data.  
\bibliography{topicsum}
\bibliographystyle{acl_natbib}
\end{document}